\documentclass{egpubl}
\usepackage{eg2025}
 
\ConferencePaper        
\CGFccbync

\usepackage[T1]{fontenc}
\usepackage{dfadobe}  

\biberVersion
\BibtexOrBiblatex
\usepackage[backend=biber,bibstyle=EG,citestyle=alphabetic,backref=true]{biblatex} 
\electronicVersion
\PrintedOrElectronic
\ifpdf \usepackage[pdftex]{graphicx} \pdfcompresslevel=9
\else \usepackage[dvips]{graphicx} \fi

\usepackage{egweblnk} 

\title[NoPe-NeRF++: Local-to-Global Optimization of NeRF with No Pose Prior]%
      {NoPe-NeRF++: Local-to-Global Optimization of NeRF \\~with No Pose Prior}

\author[D. Shi et al.]
{\parbox{\textwidth}{\centering Dongbo Shi$^{\dagger,1}$\orcid{0000-0001-6456-3388}
        Shen Cao\thanks{These authors contributed equally to this work.}$^{,2}$\orcid{0009-0003-4019-1657}
        Bojian Wu$^{2}$\orcid{0009-0007-1945-8707}
        Jinhui Guo$^{2}$\orcid{0009-0004-1129-1023}
        Lubin Fan$^{\ddagger ,2}$\orcid{0000-0002-9606-947X}
        Renjie Chen\thanks{Corresponding authors}$^{,1}$\orcid{0000-0001-8395-4392}
        Ligang Liu$^{1}$\orcid{0000-0003-4352-1431}
        Jieping Ye$^{2}$\orcid{0000-0001-8662-5818}
        }
        \\
{\parbox{\textwidth}{\centering $^1$University of Science and Technology of China, $^2$Independent Researcher
       }
}
}

\usepackage{multirow} 
\usepackage{amssymb}

\usepackage{xcolor}

\begin{document}

\teaser{
    \includegraphics[width=\linewidth]{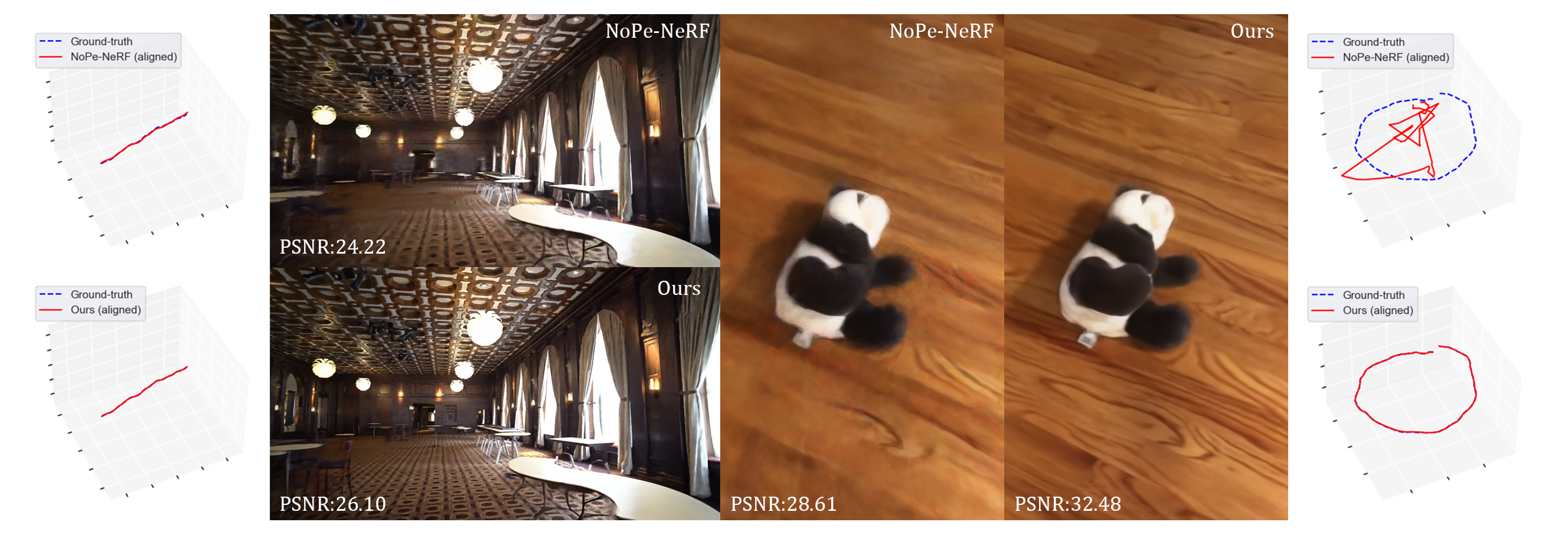}
    \centering
    \caption{In the context of unknown camera poses, accurately training high-quality NeRF while estimating the camera pose is a challenging task. The previous state-of-the-art method, NoPe-NeRF, focuses on local constraints between adjacent images and can only handle simple camera motions; it completely fails in more complex scenarios. To tackle this issue, we introduce a local-to-global training strategy, termed as NoPe-NeRF++, which considers both local position estimation and global geometric consistency, achieving not only stable estimates but also improved rendering quality and more accurate pose estimation, as demonstrated in our results.}
    \label{fig:teaser}
}

\maketitle
\begin{abstract}
In this paper, we introduce NoPe-NeRF++, a novel local-to-global optimization algorithm for training Neural Radiance Fields (NeRF) without requiring pose priors. 
Existing methods, particularly NoPe-NeRF, which focus solely on the local relationships within images, often struggle to recover accurate camera poses in complex scenarios.
To overcome the challenges, our approach begins with a relative pose initialization with explicit feature matching,  followed by a local joint optimization to enhance the pose estimation for training a more robust NeRF representation. 
This method significantly improves the quality of initial poses. 
Additionally, we introduce global optimization phase that incorporates geometric consistency constraints through bundle adjustment, which integrates feature trajectories to further refine poses and collectively boost the quality of NeRF. Notably, our method is the first work that seamlessly combines the local and global cues with NeRF, and outperforms state-of-the-art methods in both pose estimation accuracy and novel view synthesis. Extensive evaluations on benchmark datasets demonstrate our superior performance and robustness, even in challenging scenes, thus validating our design choices.


\end{abstract}  
\section{Introduction}
\label{sec:intro}

Given a set of RGB images of a 3D scene and the corresponding camera parameters, Neural Radiance Fields (NeRF)~\cite{2020NeRF} can represent the 3D scene using an implicit neural representation and synthesize high-quality images from novel viewpoints. 
Due to its strong capability for restoring geometry and appearance, NeRF has already been widely used in various tasks, such as 3D reconstruction\cite{UNISURF2021iccv, blockNeRF2022cvpr}, editing\cite{yuan2022nerf, clipnerf2022cvpr}, appearance modeling\cite{NeRFactor2021tog, zhang2022invrender}, SLAM\cite{niceslam2022cvpr, ngelslam2024icra}, etc.
Despite this, it is well-known that before training NeRF, we typically need to obtain highly accurate camera parameters for each input viewpoint, which are usually obtained from COLMAP~\cite{colmap2016}.
This approach of providing camera parameters through preprocessing presents several issues. For example, the dependency on externally supplied camera parameters limits its real scalability and applicability. 
Moreover, the poses remain fixed during training, making the introduced pose noise unavoidable and impossible to mitigate. This negatively impacts the training results, especially in scenes with substantial camera motions.
To address these issues, recent works such as NeRFmm~\cite{wang2021nerfmm}, BARF~\cite{barf2021}, SC-NeRF~\cite{SCNeRF2021}, and NoPe-NeRF~\cite{bian2022nopenerf} focus on simultaneously optimizing camera parameters and the neural implicit representation of the scene. However, these methods still make strong assumptions, including the requirement for a forward-facing scene, the ordered images, and minimal changes between adjacent frames.
Among these, NoPe-NeRF stands out as one of the most notable recent contributions. 
Similarly, without relying on any prior poses, the algorithm introduces monocular depth and relative poses to create \textit{local} loss constraints, and train a more accurate neural implicit representation of the scene. Nevertheless, NoPe-NeRF still struggles with complex scenarios, such as significant frame-to-frame camera motion changes and long-term variations.
Traditionally, to simultaneously obtain camera poses and geometry from unposed images, the Structure from Motion (SfM)~\cite{photo_tourism2006, buildrome2009, colmap2016} is used, which typically resorts to a local-to-global strategy to achieve the robust results. Basically, the local matching algorithms first roughly estimate the relative poses. The bundle adjustment is then applied to optimize and obtain a globally optimal solution. Generally, local estimation is a crucial step as it provides reasonable initial values for the subsequent global optimization. In particularly challenging cases, poor initialization can cause global optimization to fail entirely. Similarly, global optimization is also essential because relying solely on local optimization can lead to cumulative errors that cannot be corrected. Therefore, we claim that both local and global components are indispensable for this task. This is why NoPe-NeRF performs poorly in challenging cases, as it only considers local information and neglects global constraints.
To address the aforementioned problems, following the previous works, we took a step further and proposed a novel joint optimization algorithm to improve the quality of NeRF and accuracy of camera parameters, term as \textbf{NoPe-NeRF++}. Concretely, we adopted a local-to-global strategy to achieve the goal. Before that, we firstly created a maximum spanning tree to explicitly determine the association relationships among all the input images with feature matching, where the local matching pairs and the global tracks can be extracted by traversing the tree. Next, during the \textit{local joint optimization} phase, by leveraging the local matching information, we initialized the camera poses similar with NoPe-NeRF, but enhanced the estimation with random matching schemes in order to reduce accumulated errors. Additionally, inspired by bundle adjustment~\cite{monoslam2007}, we introduced a \textit{global joint optimization} to further improve the results, by combining the track trajectories and multi-view geometric consistency constraints with NeRF training, which enforces the network to reach a global optimal result.
Compared to previous methods, our algorithm achieves the state-of-the-art performance on both camera pose estimation results and the photo-realistic rendering quality of NeRF, under the same input conditions. Even with more complex camera motions, our proposed method still demonstrate the strongest robustness. The contributions can be summarized as follows. Codes and datasets will be made public upon acceptance.

\begin{itemize}
    \item We proposed a local-to-global NeRF model optimization method without pose priors, the seamless combination of the two phases demonstrate strong stability and interpretability. To the best of our knowledge, this is the first work for the task that simultaneously considers the integration of both local and global cues.
    \item Compared to NoPe-NeRF, during local optimization phase, we enhanced the estimated poses with random matching schemes to improve the stability of training.
    \item Moreover, we incorporated the global image feature-based trajectories into joint optimization to further mitigate the accumulated noises and errors.
    \item Our algorithm achieved SOTA results in both camera pose estimation and synthetic image quality across public datasets. Furthermore, extensive quantitative and qualitative analyses demonstrate that our algorithm exhibits superior robustness.
\end{itemize}

\section{Related Work}
\label{sec:related}

\subsection{Novel View Synthesis}

The goal of novel view synthesis is to generate unseen views based on the images taken from the existing views. It is usually achieved by interpolating or warping images from the neighboring point of views. Traditionally, this problem is posed as image-based rendering~\cite{chen1993view,gortler1996lumigraph,seitz1996view} or 3D geometry based rendering,
where different representations of the 3D scene include multi-plane images~\cite{MINE,single_view_mpi,stereo_magnification}, planes~\cite{automatci_photo_pop_up,tour_into_the_picture}, point clouds~\cite{point_differentiable,point_nfs} and mesh~\cite{worldsheet,FVS,SVS}, etc.

Recently, neural implicit representations have revolutionized this field. The pioneering work, NeRF~\cite{2020NeRF}, models a scene as radiance field parameterized by a deep neural network, and predicts volume density and view-dependent radiance based on spatial location and viewing direction. To enhance the quality and efficiency, significant advancements have been made, such as reducing aliasing effects~\cite{mipnerf2021}, boosting training efficiency~\cite{instant_nerf2022}, and enabling sparse viewpoint synthesis~\cite{infonerf2022}. Additionally, 3D Gaussian Splatting (3DGS)~\cite{3DGS2023} has dramatically sped up rendering during both the training and inference phases by using explicit Gaussian kernels to represent 3D scenes.

Despite the impressive advancements in novel view synthesis brought by NeRF and 3DGS, these methods still depend on offline COLMAP \cite{colmap2016} to obtain precise intrinsic and extrinsic camera parameters. This reliance significantly limits their practical use in real-world applications.

\begin{figure*}[t!]
    \centering
    \includegraphics[width=0.95\linewidth]{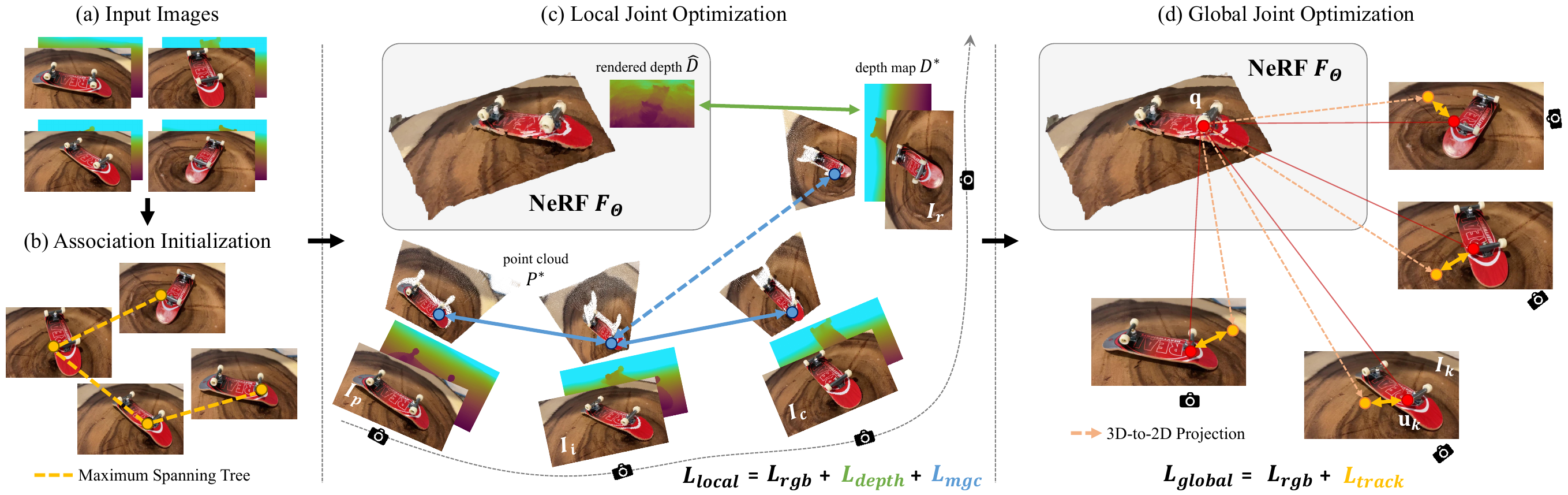}
    \caption{\textbf{Overview of NoPe-NeRF++.} Given a set of images and their corresponding camera intrinsics (a), our method first establishes the association relationships among the images using a maximum spanning tree (b). Next, we apply local optimization with photometric, depth, and multiview geometric constraints to estimate the camera poses and train the NeRF model jointly (c). Finally, these estimates are further refined globally by considering photometric differences and geometric consistency in feature matching trajectories (d).}
    \label{fig:overview}
\end{figure*}

\subsection{NeRF with Pose Optimization}

NeRF can deliver impressive scene representation when prior camera parameters are accurate. However, when this information is inaccurate or unavailable, estimating camera poses and neural representations becomes crucial. iNeRF~\cite{iNerf2021} is one of the early works that uses NeRF for pose estimation, treating it as an inverse NeRF process. Given initial camera poses, it fixes NeRF parameters and minimizes the photometric loss function to achieve its goal. Following this, BARF~\cite{barf2021} applies a coarse-to-fine strategy, yielding promising results. In addition to optimizing camera poses, NeRFmm~\cite{wang2021nerfmm} also estimates camera intrinsics during the optimization process. However, both BARF and NeRFmm are limited to achieving satisfactory optimization results only within a small range of motion, which restricts their broader applicability. 
L2G-NeRF~\cite{l2gnerf2023cvpr} employs a local-to-global registration strategy to relax the accuracy requirements of input poses during initialization. 
It calculates the local pose for each image sample and uses least squares to determine the globally optimal poses from each viewpoint.
LU-NeRF~\cite{cheng2023lunerf} initially applies BARF to a selected set of 3-5 frames to develop local models with relative poses, followed by an offline pose synchronization process for global alignment.
SAMURAI~\cite{boss2022-samurai} and SHINOBI~\cite{engelhardt2023-shinobi} are also joint optimizations of NeRF and poses that focus on improving BARF in challenging cases.
LocalRF~\cite{meuleman2023localrf} provides a robust solution for large-scale scenes from a single casually captured video, by progressively optimizing multiple overlapping local NeRF and poses.
CamP~\cite{Park2023Camp} and INN~\cite{eccv2024invertibleneuralwarpnerf} improve the performance of NeRF with pose estimation by changing the representations of camera parameters in NeRF.
To incorporate additional global constraints, TrackNeRF~\cite{tracknerf2024eccv} simultaneously optimizes sparse NeRF and camera poses based on feature tracks, achieving promising results in noisy and sparse views. 

To tackle the challenges of NeRF training when camera poses are unknown, SC-NeRF~\cite{SCNeRF2021} introduces a projected ray distance loss based on correspondences, optimizing it alongside photometric errors. This approach allows SC-NeRF to learn scene representations while simultaneously self-calibrating camera intrinsic and extrinsic parameters, as well as distortion coefficients, from scratch. However, SC-NeRF relies on pair matches to construct the projected ray distance, resulting in a lack of global constraints.
NoPe-NeRF~\cite{bian2022nopenerf} leverages monocular depth estimation priors provided by DPT~\cite{dpt2021iccv}, using undistorted depth constraints to define the relative poses between consecutive frames. This enables the joint optimization of camera parameters and NeRF.
Combined with 3D Gaussian Splatting, CF-3DGS~\cite{CF-3DGS-2024} processes input frames sequentially, progressively constructing the 3D Gaussian set by analyzing one frame at a time, where the camera parameters and 3D Gaussians are jointly optimized. This method achieves realistic scene representation and precise camera pose estimation under large camera motion changes.
However, both NoPe-NeRF and CF-3DGS lack support for processing arbitrary unordered images, limiting their broader applications.
\section{Preliminaries}
\label{sec:methods-Preliminaries}

Neural Radiance Fields (NeRF)~\cite{2020NeRF} have been proven to be an effective representation to learn dense 3D geometry of a scene, which use a neural network $F_{\Theta}:(\mathbf{x},\mathbf{d})\to(\mathbf{c},\sigma)$ to map a 3D point $\mathbf{x}\in \mathbb{R}^{3}$ and a viewing direction $\mathbf{d}\in\mathbb{R}^{3}$ to a radiance color $\mathbf{c}\in\mathbb{R}^{3}$ and a volume density $\sigma$.
Later, to render images of specified camera poses, we obtain the color $\hat{C}$ by aggregating the radiance colors on camera ray $\mathbf{r}(s)=\mathbf{o}+s\mathbf{d}$ shooting from the camera position $\mathbf{o}$ along the direction $\mathbf{d}$ between near and far bound $s_{n}$ and $s_{f}$. This volume rendering equation can be formulated as:
\begin{equation}\label{eq:volume-render}
    \hat{C}(\mathbf{r})=\int_{s_{n}}^{s_{f}}T(s)\sigma(\mathbf{r}(s))\mathbf{c}(\mathbf{r}(s),\mathbf{d})ds,
\end{equation}
where $T(t)=\mathrm{exp}(-\int_{s_{n}}^{t}\sigma(\mathbf{r}(s))ds)$ is the accumulated transmittance along ray $\mathbf{r}$.
\\
Given $N$ images $\mathcal{I}=\{I_{i}\}_{i=0}^{N-1}$ and corresponding camera parameters $\Pi=\{\pi_{i}\}_{i=0}^{N-1}$, traditional NeRF is trained by minimizing the photometric error $L_{rgb}=\sum_{i=0}^{N-1}\|I_{i}-\hat{I}_{i}\|$, where $\hat{\mathcal{I}}=\{\hat{I}_{i}\}_{i=0}^{N-1}$ are the synthesised ones from above volume rendering. 
Typically, camera parameters $\Pi$ are precomputed and fixed during the whole optimizing. 
Furthermore, prior works~\cite{bian2022nopenerf,CF-3DGS-2024,SCNeRF2021} show that $\Pi$ could be estimated and learnt together with NeRF parameters at the same time, as the ray $\mathrm{r}$ is a function of camera parameters $\Pi$.
This joint optimization is formulated as:
\begin{equation}\label{eq:joint-opt}
    \Theta^{*},\Pi^{*}=\mathrm{arg}\min_{\Theta,\Pi}L_{rgb}(\hat{\mathcal{I}},\hat{\Pi}|\mathcal{I}),
\end{equation}
where $\hat{\Pi}$ denotes the learnable camera parameters of the optimization. Note that the camera parameters $\Pi=\{\pi_{i}\}_{i=0}^{N-1}$ are composed of camera extrinsics $\mathrm{T}=\{\mathrm{T}_{i}\}_{i=0}^{N-1}$ and intrinsics $\mathrm{K}$ and other distortions, where $\mathrm{T}_{i}=[\mathrm{R}_{i}|\mathrm{t}_i]$ is a rigid transformation with a rotation matrix $\mathrm{R}_{i}\in\mathrm{SO(3)}$ and a translation vector $\mathrm{t}_i\in\mathbb{R}^3$. 
we only focus on the extrinsics $\mathrm{T}$ in our work, while the other parameters are settled during the optimization.


\section{Method}\label{sec:methods}

\textbf{Overview.}
Given a set of images and their corresponding camera intrinsics, our method uses a local-to-global strategy to jointly estimate camera poses and the NeRF, as illustrated in Fig.~\ref{fig:overview}. To effectively handle large view changes at the start, we introduce an association relationships initialization. Specifically, we construct a Maximum Spanning Tree (MST) to establish adjacency relationship among all input images, where each node represents an image, and the edge weights are determined by the maximum feature matches. This approach allows us to extract local matching pairs and global tracks from the tree, which proves effective in subsequent optimization. During the local joint optimization phase, we begin with the tree's root node, as it represents the image with the most matching pixels. We then progressively determine the relative pose between adjacent nodes based on the tree's adjacency structure to complete the initialization. These initial estimations serve as the starting poses for NeRF. Throughout training, we enforce photometric and depth consistency within a single view and geometric consistency across multiple views. Importantly, when introducing multi-view geometric consistency constraints, we also introduce a random matching scheme and incorporate non-adjacent random images as supervision. This inclusion allows global information to be integrated into local optimization, reducing accumulated errors in pose estimation and yielding more stable results. In the global joint optimization phase, we further refine the camera poses and NeRF by applying global geometric consistency constraints through the tracking trajectory. This involves supervising all images within the trajectory, similar to bundle adjustment, to enhance camera poses and the NeRF on a global scale.

\subsection{Association Initialization}
\label{sec:method-assoc}

In order to achieve more stable camera pose estimation during joint optimization and to handle large view changes, we propose leveraging the explicit image feature matching relationship to re-establish the association relationships among input images. This approach takes advantage of the robustness and stability of feature matching, using the number of matched pixels as a critical indicator of image view similarity, thereby enabling accurate estimation of the relative pose transformation between image pairs.

Specifically, we construct a Maximum Spanning Tree (MST) based on feature matching pairs between images, see Fig.~\ref{fig:overview}(b). The process involves the following steps.
Firstly, we use the deep feature point extraction method, SuperPoint~\cite{detone2018superpoint}, and the deep feature matching algorithm, SuperGlue~\cite{sarlin20superglue}, to compute the feature matches among all images. We define the weight between two images as the number of matching pairs. Then we construct the MST using Kruskal's algorithm, which sorts edges by weight from highest to lowest, ensuring no loops in the tree.
By traversing this MST, we can extract the local matching pairs of images connected by an edge, and the global feature matching trajectories or track by union-find structure. These are used in subsequent local and global joint optimization to maintain geometric consistency and guide the optimization of camera relative pose estimation.

\subsection{Local Joint Optimization}
\label{sec:methods-local}

Inspired by NoPe-NeRF~\cite{bian2022nopenerf}, we adopt a similar joint optimization strategy to simultaneously estimate camera poses and train the NeRF. For constraints on a single view, our approach considers both photometric and monocular depth loss. 
In addition, geometric constraints are introduced to improve the robustness. However, unlike NoPe-NeRF, it relies on the construction of dense point clouds and uses the chamfer distance to measure discrepancies, while this loss term struggles to capture the relative transformation accurately when camera motions are complex. In our procedure, we employ explicit feature matching to ensure robust geometric consistency across multiple associated views, as in Eq. \ref{eq:loss-pc} and \ref{eq:loss-mgc}, which is primarily derived from adjacent nodes obtained through the pre-constructed MST, focusing on local consistency.
Therefore, we refer to this process as local joint optimization.

\textbf{Mono-depth distortion constraint.}
As evidenced by previous works, monocular depth from the input image can provide a strong geometric prior for NeRF training. It helps reduce shape ambiguity in addition to the photometric constraint. In our approach, we leverage the estimated depth $D$ from DPT~\cite{dpt2021iccv} for supervision. However, the absolute scale of depth map is unknown. To address this, we apply two linear transformations to map the depth to the real scale. Specifically, we define the transformation as $\Psi=\{\alpha_i, \beta_i\}$, where $\alpha$ and $\beta$ denote a scale and a shift, respectively. With the transformed depth map $D^{*}=\alpha D + \beta$, we integrate this additional depth supervision into our optimization process,
\begin{equation}\label{eq:loss-depth}
L_{depth}=\sum_{i=0}^{N-1}\|D_{i}^{*}-\hat{D}_{i}\|,
\end{equation}
where $D_{i}^{*}$ and $\hat{D}_{i}$ denote the transformed depth map and the NeRF rendered depth map of image $I_i$, respectively.

\textbf{Multiview geometric consistency constraint.}
Since the camera poses can be estimated by multi-view geometry, we also incorporate this constraint into our joint optimization framework. 
Taking two images $I_i$ and $I_j$ as an example, we back-project the corresponding depth maps to obtain the point sets $P_i^{*}$ and $P_j^{*}$. We then measure the geometric consistency as follows,
\begin{equation}\label{eq:loss-pc}
l_{gc}(P_i^{*},P_j^{*}) = \sum_{k} \| \mathbf{p}_{j,k}^{*} - T_{i,j} \cdot \mathbf{p}_{i,k}^{*} \|,
\end{equation}
where $T_{i,j}$ denotes the relative pose that transforms from $P_i^{*}$ to $P_j^{*}$, $\mathbf{p}_{i,k}^{*}$ and $\mathbf{p}_{j,k}^{*}$ represent the corresponding matching point pair in $P_i^{*}$ and $P_j^{*}$, respectively, where the correspondence originates from pixel level matching. Concretely, for each input image $I_i$, we take its parent node $I_p$ and child node $I_c$ from the MST as references for multi-view geometric consistency. Since this constraint is local, it may be affected by noise and accumulated errors. Therefore, we incorporate a random matching scheme, which randomly introduces another non-adjacent image $I_r$ with matching points for constraint. The matching number is at least 100 pairs for $I_r$ to be considered for random selection. Since the image $I_r$ is randomly selected, it can be regarded as integrating a weak global information into the constraint, which strengthens the constraint of multi-view geometric consistency. Therefore, the loss function can be defined as:
\begin{equation}\label{eq:loss-mgc}
L_{mgc} = \sum_{i} l_{gc}(P_i^{*}, P_{p}^{*}) + l_{gc}(P_i^{*}, P_{c}^{*}) + l_{gc}(P_i^{*}, P_{r}^{*}).
\end{equation}

\textbf{Optimization formulation.}
We obtain the loss function of our local joint optimization as:
\begin{equation}\label{eq:Loss-local}
    L_{local}=L_{rgb}+\lambda_{depth} L_{depth}+\lambda_{mgc} L_{mgc},
\end{equation}
where $\lambda_{depth}$ and $\lambda_{mgc}$ are the weighting factors for corresponding loss terms.
Then the local joint model can be formulated as:
\begin{equation}\label{eq:model-local}
    \Theta^{*},\Pi^{*}, \Psi^{*}=\mathrm{arg}\min_{\Theta,\Pi,\Psi}L_{local}(\hat{\mathcal{I}},\hat{\Pi},\hat{\Psi}|\mathcal{I},D).
\end{equation}

\textbf{Initialization of the optimization.}  
Since NeRF training relies on camera poses, and to enhance joint optimization, some methods like \cite{iNerf2021,wang2021nerfmm,barf2021} require initial poses that are very close to the ground truth or derived from COLMAP. Meanwhile, approaches like InstantSplat~\cite{fan2024instantsplat} employ 3D reconstruction (DUSt3R~\cite{dust3r_cvpr24}) to initialize camera parameters. A better initialization consistently yields better results. However, we choose a rough initialization using monocular depth which is regarded as geometric prior in our local optimization. Experiments prove that this simple initialization is sufficient for delivering stable results.
First, we use the monocular depth map $D_i$ with the default scale and the camera intrinsics to back-project and restore the point cloud $P_i^{*}$ for each image $I_i$. Then we use the association relationships from the MST to estimate the relative pose transformation $T_{i,j}$ for each child node $I_j$ in the tree. The relative rigid transformation $T_{i,j}$ is obtained by minimizing the distance between $P_j^{*}$ and $T_{i,j} \cdot P_i^{*}$. We treat this as a point cloud registration problem and solve it using the Iterative Closest Point (ICP) algorithm. 

\subsection{Global Joint Optimization}
\label{sec:method-global}

After local joint optimization, we obtain a set of camera poses that roughly align with the ground truth, as evident from Fig.~\ref{fig:pose-co3dv2}.
However, several issues remain. While the introduction of the mono-depth provides a strong geometric prior for NeRF training and helps reduce shape ambiguity that would otherwise rely solely on photometric constraints, the estimated monocular depth is often inaccurate, hindering optimal NeRF training. Additionally, the geometric consistency constraint is only applied to limited number of adjacent images, leading to accumulated errors that further impact the NeRF training. To address this, we aim to further improve the accuracy of camera pose estimation and the performance of NeRF by focusing on two main aspects: reducing the camera's accumulated errors and providing more accurate geometric priors for NeRF training. Inspired by the bundle adjustment algorithm, we introduce global geometric consistency constraints to achieve these goals.

Specifically, we use image matching and tracking in the association initialization phase to add geometric consistency constraints for more images. 
A track consists of a set of corresponding feature-matched pixels across a sequence of images, all representing the same 3D point in the scene. Using the camera poses obtained through local joint optimization, we can calculate the reprojection error. This error measures the distance between the reprojection of the point $\mathbf{q}$ and the corresponding image feature pixel $\mathbf{u}$, allowing us to evaluate the accuracy of the camera pose.
\begin{equation}
l_{proj}(\mathbf{q}) = \sum_{k} \| \mathbf{u}_k - K \cdot T_k \cdot \mathbf{q} \|,
\end{equation}
where $K$ is the camera intrinsic parameter, $T_k$ is the camera pose of the $k$-th image in the track, and $\mathbf{u}_k$ represents the pixel coordinates of the feature point in the $k$-th image. 
It is worth noting that for the initialization of $\mathbf{q}$, we randomly select a pixel from one frame of a track, such as pixel $\mathbf{u}_k$ from the $k$-th image. Using the initial NeRF obtained from the local optimization phase, we render the depth value of this pixel and calculate the corresponding 3D coordinates. This 3D point is then treated as a 3D optimization variable and further optimized during the global optimization phase.
Finally, we define the reprojection error calculated for all tracks within the association relationships of the MST as the track constraint:
\begin{equation}
    \label{eq:loss-track}
    L_{track}= \sum_{\mathbf{q} \in Q} l_{proj}(\mathbf{q}),
\end{equation}
where $Q$ is the set of tracking 3D points, whose average size is 10k in our experiments.

Compared to the multiview geometric consistency constraint in the local joint optimization, the track constraint simultaneously constrains a set of image poses across all trajectories. This results in a broader impact range than $L_{mgc}$ and is beneficial for reducing accumulated errors and obtaining stable results. Additionally, for each track, the same 3D point is used as the reference and further optimized, leading to more accurate geometric priors. Consequently, $L_{track}$ serves as a stronger global geometric constraint.

Then the overall loss function of our global joint optimization is defined as:
\begin{equation}\label{eq:Loss-global}
    L_{global}=L_{rgb}+\lambda_{track} L_{track},
\end{equation}
where $\lambda_{track}$ is the weighting factor.
Our global joint model can be formulated as:
\begin{equation}
    \label{eq:model-global}
    \Theta^{*},\Pi^{*},Q^{*}=\mathrm{arg}\min_{\Theta,\Pi,Q}L_{global}(\hat{\mathcal{I}},\hat{\Pi},\hat{Q}|\mathcal{I}).
\end{equation}
\section{Experiments}
\label{sec:exps}

\subsection{Experimental Setup}

\begin{figure*}[p!]
    \centering
    \includegraphics[width=.95\linewidth]{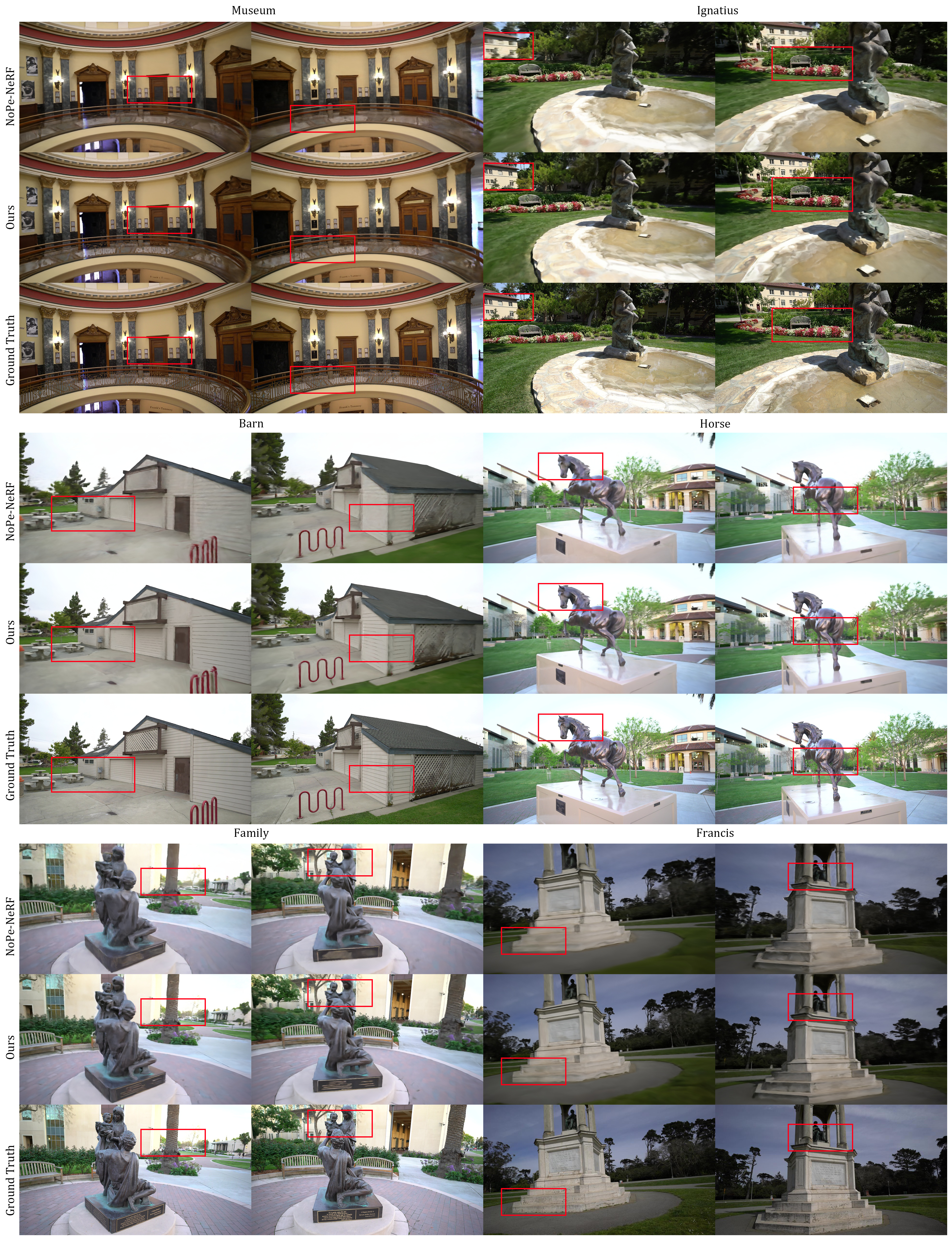}
    \caption{\textbf{Qualitative comparison for novel view synthesis on Tanks and Temples.} Our method can convey richer details, while NoPe-NeRF appears more blurred. The red boxes indicate the regions to focus on.}
    \label{fig:render-T&T}
\end{figure*}

\textbf{Datasets}. We conduct experiments on two benchmark datasets, \emph{Tanks and Temples}~\cite{Tanks_Temples} and \emph{CO3D V2}~\cite{co3d}, to evaluate our performance in novel view synthesis and pose estimation. Specifically, for \textbf{Tanks and Temples}, we create two categories of scenes. The first is directly adapted from NoPe-NeRF~\cite{bian2022nopenerf} and includes 8 indoor and outdoor scenes with maximum relative rotation angles between any two frames ranging from 30$^{\circ}$ to 70$^{\circ}$. The second, more challenging dataset is created by us from the original videos of \cite{Tanks_Temples}. We choose \emph{'Family'} video as our experimental scene and randomly extract 100 images from it. This results in significant transformations between adjacent frames, allowing us to demonstrate the stability of our method in non-sequential cases. For \textbf{CO3D V2}, borrowed from CF-3DGS\cite{CF-3DGS-2024}, we select 5 scenes of different categories of objects, all involving large camera motions. We follow their experimental setup on these object-centric scenes to show the robustness of our method. For all mentioned scenes, we sample 1 from every 8 frames for novel view synthesis, and the rest of the images are used for training.

\textbf{Metrics.} We use standard evaluation metrics, including PSNR, SSIM~\cite{SSIM} and LPIPS~\cite{LPIPS} to evaluate the quality of novel view synthesis. For pose estimation, we rely on the Absolute Trajectory Error (ATE) and Relative Pose Error (RPE)~\cite{barf2021,bian2022nopenerf,CF-3DGS-2024}. $\mathrm{RPE}_r$ and $\mathrm{RPE}_t$ are utilized to measure the accuracy of rotation and translation, respectively. To ensure the metrics are comparable on the same scale, we align the camera poses by Umeyama's method~\cite{Umeyama} for both estimation and evaluation. Since all the mentioned datasets lack ground truth camera poses, we use COLMAP~\cite{colmap2016} to estimate the poses as reference.

\textbf{Implementation details.}
Our implementation is primarily based on NoPe-NeRF~\cite{bian2022nopenerf}.
We optimize the mono-depth transformation ($\Psi$) parameters only during the local optimization phase, and track points ($Q$) only during global phase. NeRF ($F_{\Theta}$) and camera poses ($\Pi$) parameters are refined in both phases. 
Moreover, each pose is formatted as a combination of an axis-angle representation $\mathrm{q}\in\mathrm{SO(3)}$ and a translation vector $\mathrm{t}\in\mathbb{R}^3$.
The initial learning rate for NeRF is set at 0.001 and the rates for other parameters are set at 0.0005.
During the local phase, we train the joint optimization (Eq.~\ref{eq:model-local}) for 40k steps with $\lambda_{depth}=0.04$, $\lambda_{mgc}=1.0$ of loss terms in Eq.~\ref{eq:Loss-local}.
For global optimization, the model (Eq.~\ref{eq:model-global}) is first trained for 10k steps with $\lambda_{track}=10.0$ in Eq.~\ref{eq:Loss-global}, followed by 5k steps with $\lambda_{track}=1.0$. For other training parameters, we use the same settings as NoPe-NeRF, such as learning rate decreasing strategy. 

\begin{figure}[!t]
    \centering
    \includegraphics[width=\linewidth]{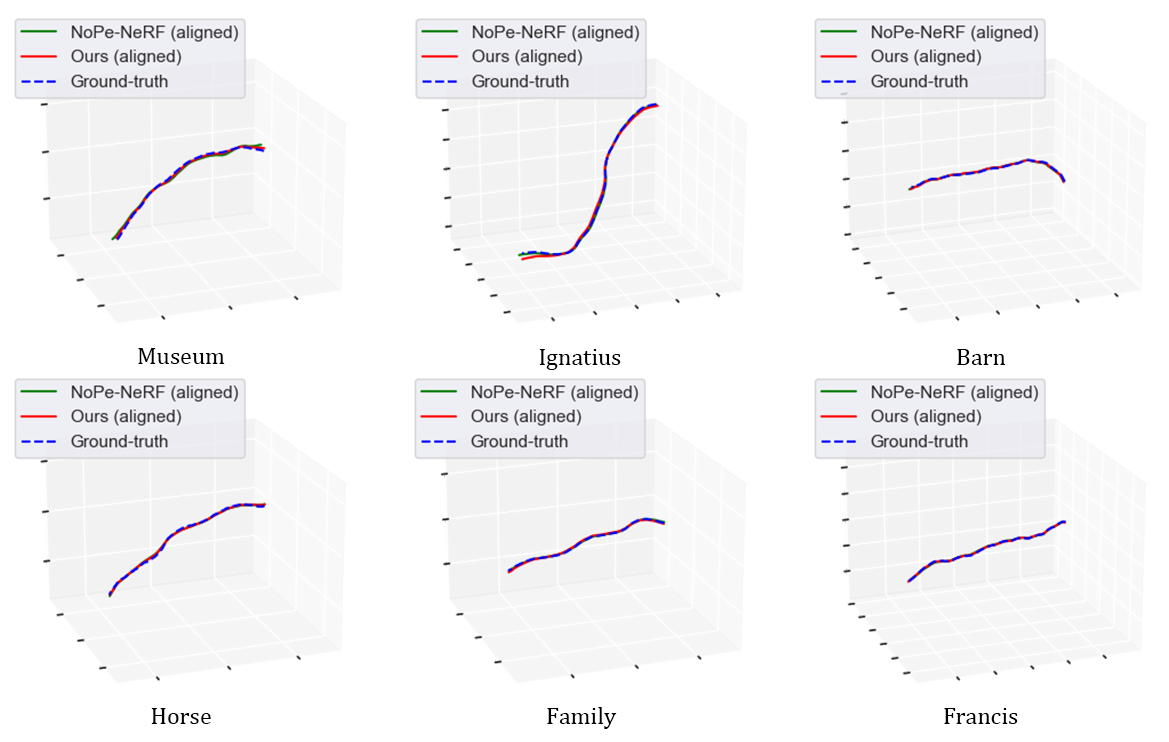}
    \caption{\textbf{Comparison of pose estimation on Tanks and Temples.}}
    \label{fig:pose-T&T}
\end{figure}

\subsection{Comparing with Pose-Unknown Methods}
We compare our method with state-of-the-art pose-unknown methods, including NoPe-NeRF~\cite{bian2022nopenerf}, BARF~\cite{barf2021}, and CF-3DGS~\cite{CF-3DGS-2024}. Notably, for novel view synthesis, following NeRFmm~\cite{wang2021nerfmm}, we obtain the camera poses of test views by minimizing the photometric errors between the synthesized and test images, while keeping the trained NeRF parameters fixed. The test camera poses are initialized to the poses of most geometry-related training images to expedite convergence. All baselines use the same pre-processing method to ensure a fair comparison.

\subsubsection{With NeRF-based Methods}
We first compare our results with NoPe-NeRF and BARF, both of which use the NeRF model as the scene representation. 

\begin{table}[!t]
    \centering
    \resizebox{\linewidth}{!}{
    \begin{tabular}{c cccc cccc ccc} \hline
    \multirow{2}{*}{Scenes} & \multicolumn{3}{c}{Ours} & & \multicolumn{3}{c}{NoPe-NeRF} & & \multicolumn{3}{c}{BARF}\\ \cline{2-4}\cline{6-8}\cline{10-12}
                            & PSNR$\uparrow$ & SSIM$\uparrow$ & LPIPS$\downarrow$ & & PSNR & SSIM & LPIPS & & PSNR & SSIM & LPIPS \\ \hline
    Church                  & 25.08 & \textbf{0.76} & \textbf{0.33}    & & \textbf{25.17} & 0.73 & 0.39    & & 23.17 & 0.62 & 0.52   \\
    Barn                    & \textbf{26.69} & \textbf{0.73} & \textbf{0.39}    & & 26.35 & 0.69 & 0.44    & & 25.28 & 0.64 & 0.48   \\
    Museum                  & \textbf{27.55}& \textbf{0.81}& \textbf{0.29} & & 26.77 & 0.76 & 0.35    & & 23.58 & 0.61 & 0.55   \\
    Family                  & \textbf{26.66}& \textbf{0.78}& \textbf{0.36} & & 26.01 & 0.74 & 0.41    & & 23.04 & 0.61 & 0.56   \\
    Horse                   & 27.08& \textbf{0.84}& \textbf{0.25} & & \textbf{27.64} & \textbf{0.84} & 0.26    & & 24.09 & 0.72 & 0.41   \\
    Ballroom                & \textbf{26.65}& \textbf{0.81}& \textbf{0.27} & & 25.33 & 0.72 & 0.38    & & 20.66 & 0.50 & 0.60   \\
    Francis                 & \textbf{29.92}& \textbf{0.82}& \textbf{0.36} & & 29.48 & 0.80 & 0.38    & & 25.85 & 0.69 & 0.57   \\
    Ignatius                & \textbf{24.33 }& \textbf{0.63 }& \textbf{0.44}    & & 23.96 & 0.61 & 0.47    & & 21.78 & 0.47 & 0.60   \\ \hline
    mean                    & \textbf{26.74}& \textbf{0.77}& \textbf{0.34}& & 26.34 & 0.74 & 0.39    & & 23.42 & 0.61 & 0.54    \\ \hline
    \end{tabular}
    }
    \caption{\textbf{Quantitative comparison of novel view synthesis on Tanks and Temples.} Each baseline method is trained with its public code under the original settings and evaluated with the same evaluation protocol. The best results are highlighted in bold.}
    \label{table:render-T&T}
\end{table}

\begin{table}[!t]
    \centering
    \resizebox{\linewidth}{!}{
    \begin{tabular}{c cccc cccc ccc} \hline
    \multirow{2}{*}{Scenes} & \multicolumn{3}{c}{Ours} & & \multicolumn{3}{c}{NoPe-NeRF}  & & \multicolumn{3}{c}{BARF} \\ \cline{2-4}\cline{6-8}\cline{10-12}
                            & $\mathrm{RPE}_{t}\downarrow$ & $\mathrm{RPE}_{r}\downarrow$ & ATE$\downarrow$ & & $\mathrm{RPE}_{t}$ & $\mathrm{RPE}_{r}$ & ATE & & $\mathrm{RPE}_{t}$ & $\mathrm{RPE}_{r}$ & ATE \\ \hline
    Church                  & \textbf{0.012} & 0.011 & \textbf{0.001}    & & 0.034 & \textbf{0.008} & 0.008           & & 0.114 & 0.038 & 0.052\\
    Barn                    & \textbf{0.010} & \textbf{0.024} & \textbf{0.001}    & & 0.046 & 0.032 & 0.004                    & & 0.314 & 0.265 & 0.050\\
    Museum                  & \textbf{0.032} & \textbf{0.133} & \textbf{0.002}    & & 0.207 & 0.202 & 0.020                    & & 3.442 & 1.128 & 0.263\\
    Family                  & \textbf{0.023} & 0.029 & 0.002    & & 0.047 & \textbf{0.015} & \textbf{0.001}  & & 1.371 & 0.591 & 0.115\\
    Horse                   & \textbf{0.086} & 0.050 & \textbf{0.002}    & & 0.179 & \textbf{0.017} & 0.003           & & 1.333 & 0.394 & 0.014\\
    Ballroom                & \textbf{0.019} & \textbf{0.015} & \textbf{0.001}    & & 0.041 & 0.018 & 0.002                    & & 0.531 & 0.228 & 0.018\\
    Francis                 & \textbf{0.012} & 0.036 & \textbf{0.001}    & & 0.057 & \textbf{0.009} & 0.005           & & 1.321 & 0.558 & 0.082\\
    Ignatius                & \textbf{0.024} & 0.032 & 0.003    & & 0.026 & \textbf{0.005} & \textbf{0.002}  & & 0.736 & 0.324 & 0.029\\ \hline
    mean                    & \textbf{0.027} & 0.041 & \textbf{0.002}    & & 0.080 & \textbf{0.038} & 0.006  & & 1.046 & 0.441 & 0.078\\ \hline
    \end{tabular}
    }
    \caption{\textbf{Quantitative comparison of pose accuracy on Tanks and Temples.} Note that, we use COLMAP poses as the ground truth. The unit of $RPE_{r}$ is in degrees, ATE is in the ground truth scale and $RPE_{t}$ is scaled by 100. The best results are highlighted in bold.}
    \label{table:pose-T&T}
\end{table}

We first present the qualitative comparisons of novel view synthesis and pose estimation on Tanks and Temples in Fig.~\ref{fig:render-T&T} and Fig.~\ref{fig:pose-T&T}. Notably, our method obviously enhances the rendering quality, as highlighted in the image. Our results convey richer details, while NoPe-NeRF appears more blurred. This improvement is attributed to the accurate camera poses aligned with ground truth, which is also directly validated in Fig.~\ref{fig:pose-T&T}. Following this, we summarize the numerical accuracy in Tab.~\ref{table:render-T&T} and Tab.~\ref{table:pose-T&T}. On average, our method outperforms existing state-of-the-art approaches across all metrics.

\begin{figure*}[!t]
    \centering
    \includegraphics[width=.9\linewidth]{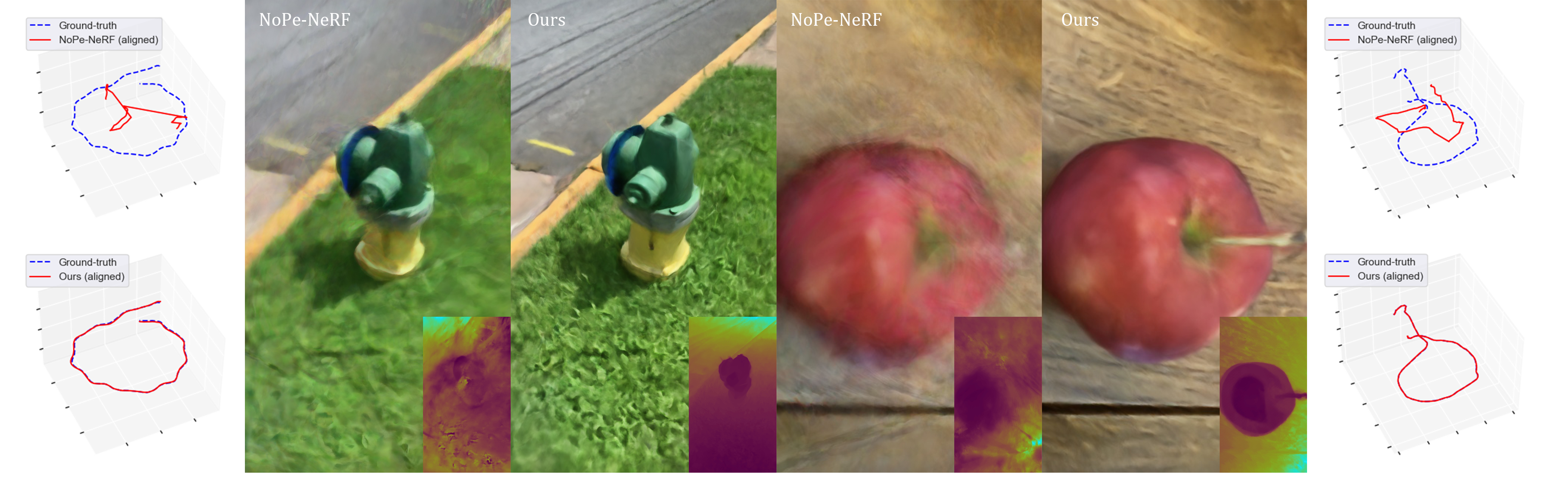}
    \caption{\textbf{Qualitative comparison of novel view synthesis on CO3D V2.} We also visualize the rendered depth maps of NeRF models (bottom right of each image) for both methods. Notice that our geometric quality is much better than NoPe-NeRF.}
    \label{fig:render-co3dv2}
\end{figure*}

\begin{table*}[!t]
    \centering
    \resizebox{\textwidth}{!}
    {
    \begin{tabular}{l cccc cccc cccc cccc ccc} \hline
    \multirow{2}{*}{Methods} & \multicolumn{3}{c}{110\_13051\_23361} & & \multicolumn{3}{c}{415\_57112\_110099} & & \multicolumn{3}{c}{106\_12648\_23157} & & \multicolumn{3}{c}{245\_26182\_52130} & & \multicolumn{3}{c}{34\_1403\_4393} \\ \cline{2-4}\cline{6-8}\cline{10-12}\cline{14-16}\cline{18-20}
                & PSNR$\uparrow$ & SSIM$\uparrow$ & LPIPS$\downarrow$   & &  PSNR & SSIM & LPIPS   & &  PSNR & SSIM & LPIPS  & &  PSNR & SSIM & LPIPS   & &  PSNR & SSIM & LPIPS  \\   \hline
    NoPe-NeRF   & 26.86 & 0.73 & 0.47  & & 24.78 & \textbf{0.64} & 0.55  & & 20.41 & 0.46 & 0.58  & & 25.05 & 0.80 & 0.49  & & 28.62 & 0.80 & 0.35\\
    Ours        & \textbf{30.26} & \textbf{0.83} & \textbf{0.31}  & & \textbf{25.40} & \textbf{0.64} & \textbf{0.53}  & & \textbf{20.91} & \textbf{0.48} & \textbf{0.52}  & & \textbf{27.18} & \textbf{0.83} & \textbf{0.43}  & & \textbf{32.00} & \textbf{0.87} & \textbf{0.29}\\ \hline
    \end{tabular}
    }
    \caption{\textbf{Comparison of novel view synthesis on CO3D V2.}}
    \label{table:render-co3dv2}
\end{table*}

\begin{table*}[!t]
    \centering
    \resizebox{\textwidth}{!}
    {
    \begin{tabular}{l cccc cccc cccc cccc ccc} \hline
    \multirow{2}{*}{Methods} & \multicolumn{3}{c}{110\_13051\_23361} & & \multicolumn{3}{c}{415\_57112\_110099} & & \multicolumn{3}{c}{106\_12648\_23157} & & \multicolumn{3}{c}{245\_26182\_52130} & & \multicolumn{3}{c}{34\_1403\_4393} \\ \cline{2-4}\cline{6-8}\cline{10-12}\cline{14-16}\cline{18-20}
                & $\mathrm{RPE}_{t}\downarrow$ & $\mathrm{RPE}_{r}\downarrow$ & ATE$\downarrow$   & & $\mathrm{RPE}_{t}$ & $\mathrm{RPE}_{r}$ & ATE  & & $\mathrm{RPE}_{t}$ & $\mathrm{RPE}_{r}$ & ATE & & $\mathrm{RPE}_{t}$ & $\mathrm{RPE}_{r}$ & ATE  & & $\mathrm{RPE}_{t}$ & $\mathrm{RPE}_{r}$ & ATE  \\   \hline
    NoPe-NeRF   & 0.400 & 1.966 & 0.046  & & 0.326 & 1.919 & 0.054  & & 0.387 & 1.312 & 0.049  & & 0.587 & 1.867 & 0.038  & & 0.591 & 1.313 & 0.053\\
    Ours        & \textbf{0.015} & \textbf{0.065} & \textbf{0.001}  & & \textbf{0.004} & \textbf{0.020} & \textbf{0.001}  & & \textbf{0.010} & \textbf{0.116} & \textbf{0.001}  & & \textbf{0.011} & \textbf{0.059} & \textbf{0.001}  & & \textbf{0.030} & \textbf{0.110} & \textbf{0.001}\\ \hline
    \end{tabular}
    }
    \caption{\textbf{Comparison of pose estimation on CO3D V2.}}
    \label{table:pose-co3dv2}
\end{table*}

Notably, in terms of pose accuracy, our method performs better in $\mathrm{RPE}_{t}$ and $\mathrm{ATE}$, but is slightly worse on $\mathrm{RPE}_{r}$. We claim that the calculation for $\mathrm{RPE}_{r}$ is directly related to the accuracy of adjacent frame poses, which is where local optimization excels; hence, NoPe-NeRF achieves better results there. However, we focus more on global optimization, leading to higher performance in the other two metrics. Additionally, improved global consistency contributes to higher rendering quality, as demonstrated above.

\begin{figure}[!t]
    \centering
    \includegraphics[width=\linewidth]{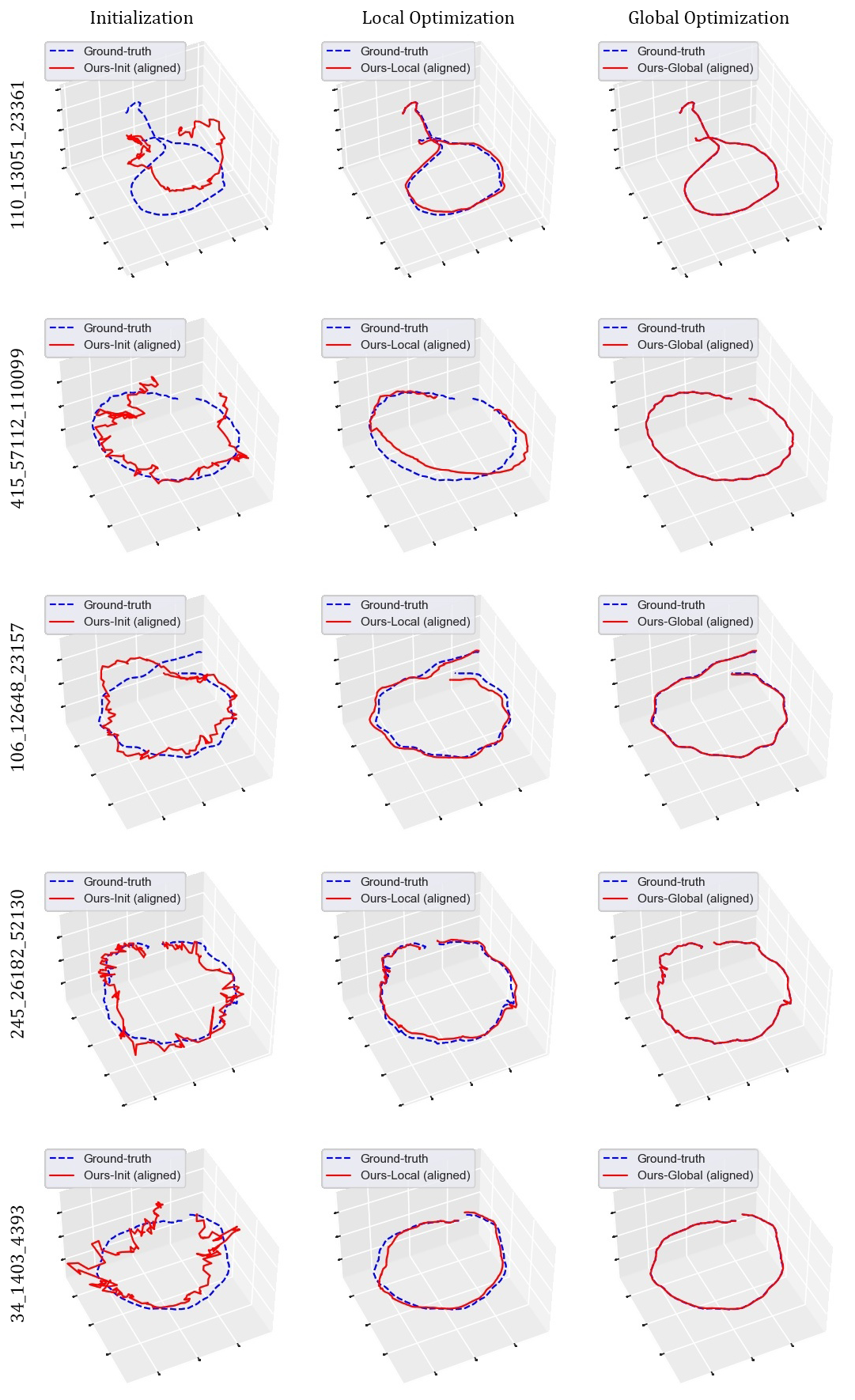}
    \caption{\textbf{Pose estimation results for each phase on CO3D V2.}}
    \label{fig:pose-co3dv2}
\end{figure}

Notice that, in the Tanks and Temples datasets, the camera movements are relatively small, and estimating relative poses between adjacent frames during the local phase is quite easy. To better highlight the advantages of our algorithm, we choose the more challenging CO3D V2 dataset for validation, where camera pose changes are more obvious and the range of camera movements between frames is larger. Local pose estimation struggles to achieve good results in this case. As shown in Fig.~\ref{fig:render-co3dv2}, Tab.~\ref{table:render-co3dv2} and \ref{table:pose-co3dv2}, our algorithm still delivers better rendering quality and more accurate camera poses, while NoPe-NeRF may completely fail. In Fig.~\ref{fig:render-co3dv2}, we also present a comparison of depth maps rendered by the trained NeRF, illustrating that our improved geometry results from optimizing both NeRF parameters and track points during the global optimization phase, allowing us to learn better geometry.

On this more challenging dataset, we further analyze the performance of our algorithm at each phase, as shown in Fig.~\ref{fig:pose-co3dv2}. It is clear that during the initialization, we can only obtain a very rough estimate of the camera poses. However, in the local optimization phase, observing that the relationships between adjacent frames are well constrained, allows the optimized poses to get closer to the ground truth, but the noticeable accumulated errors are still present, as seen in the examples of the second and fourth rows. After applying global optimization, the final poses converge accurately.

In terms of training time, our method requires approximately 30 hours on average for a single scene, compared to 12 hours for Vanilla-NeRF and 20 hours for NoPe-NeRF. This total includes about 20 hours for the local optimization phase and around 10 hours for the global refinement. The increased training time is primarily due to the use of NeRF as a back-end. We acknowledge this trade-off in time efficiency for significantly improved accuracy and robustness, which we believe is worthwhile.

\begin{table}[!t]
    \centering
    \resizebox{\linewidth}{!}{
    \begin{tabular}{c c| ccc c ccc}\hline
        & \multirow{2}{*}{scenes}  & \multicolumn{3}{c}{Ours} & & \multicolumn{3}{c}{CF-3DGS}\\ \cline{3-5}\cline{7-9}
        &                          & $RPE_{t}\downarrow$ & $RPE_{r}\downarrow$ & ATE$\downarrow$ & & $RPE_{t}$ & $RPE_{r}$ & ATE \\ 
        \hline
    \multirow{8}{*}{\rotatebox{90}{Tanks and Temples}}         & Church                  & 0.012 & \textbf{0.011} & \textbf{0.001}    &  & \textbf{0.008} & 0.018 & 0.002    \\
        & Barn                    & \textbf{0.010} & \textbf{0.024} & \textbf{0.001}    & & 0.034 & 0.034 & 0.003    \\
        & Museum                  & \textbf{0.032} & \textbf{0.133} & \textbf{0.002}    & & 0.052 & 0.215 & 0.005    \\
        & Family                  & 0.023 & 0.029 & \textbf{0.002}  & & \textbf{0.022} & \textbf{0.024} & \textbf{0.002}    \\
        & Horse                   & \textbf{0.086} & \textbf{0.050} & \textbf{0.002}    & & 0.122 & 0.057 & 0.003    \\
        & Ballroom                & \textbf{0.019} & \textbf{0.015} & \textbf{0.001 }   & & 0.037 & 0.024 & 0.003    \\
        & Francis                 & \textbf{0.012} & \textbf{0.036} & \textbf{0.001}    & & 0.029 & 0.154 & 0.006   \\
        & Ignatius                & \textbf{0.024} & \textbf{0.032} & \textbf{0.003}    & & 0.033 & \textbf{0.032} & 0.005    \\ \hline
    \multirow{5}{*}{\rotatebox{90}{CO3D V2}}          & {110\_13051\_23361}     & \textbf{0.015} &  \textbf{0.065} &   \textbf{0.001}  & & 0.061 & 0.253 & 0.005  \\
        & {415\_57112\_110099}    & \textbf{0.004} &  \textbf{0.020} &   \textbf{0.001}  & & 0.149 & 1.412 & 0.016  \\
        & {106\_12648\_23157}     & \textbf{0.010} &  \textbf{0.116} &   \textbf{0.001}  & & 0.094 & 0.360 & 0.008  \\
        & {245\_26182\_52130}     & \textbf{0.011} &  \textbf{0.059} &   \textbf{0.001}  & & 0.056 & 0.324 & 0.006  \\
        & {34\_1403\_4393}        & \textbf{0.030} & \textbf{0.110} & \textbf{0.001}  & & 0.108 & 0.390 & 0.010  \\ 
        \hline
    \end{tabular}
    }
    \caption{\textbf{Comparison to CF-3DGS on pose accuracy.} The best results are highlighted in bold.}
    \label{table:pose&render-cf3dgs}
\end{table}

\subsubsection{With CF-3DGS}
In Tab.~\ref{table:pose&render-cf3dgs}, we compare pose estimation accuracy with CF-3DGS. Overall, our method significantly outperforms CF-3DGS across all test datasets. 
Notably, while CF-3DGS also considers local information during initialization, we introduce a random matching scheme during the local optimization phase to incorporate weak global information which results in better initialization. In other words, although CF-3DGS also utilizes global optimization, in certain challenging scenarios, relatively poor initial values for camera poses may not be fully corrected through global optimization.
This example underscores the interplay between local and global optimization; their collaboration is crucial for achieving optimal results. Additionally, we do not compare the rendering quality here, as it is well-known that the rendering quality of the CF-3DGS base model is obviously better than that of NeRF. We focus solely on pose accuracy to validate the rationale of our algorithm. We firmly believe that using 3D Gaussian Splatting as the base model would yield even better rendering results in our case.

\subsubsection{Results on More Challenging Scene}
\begin{figure}[!t]
    \centering
    \includegraphics[width=\linewidth]{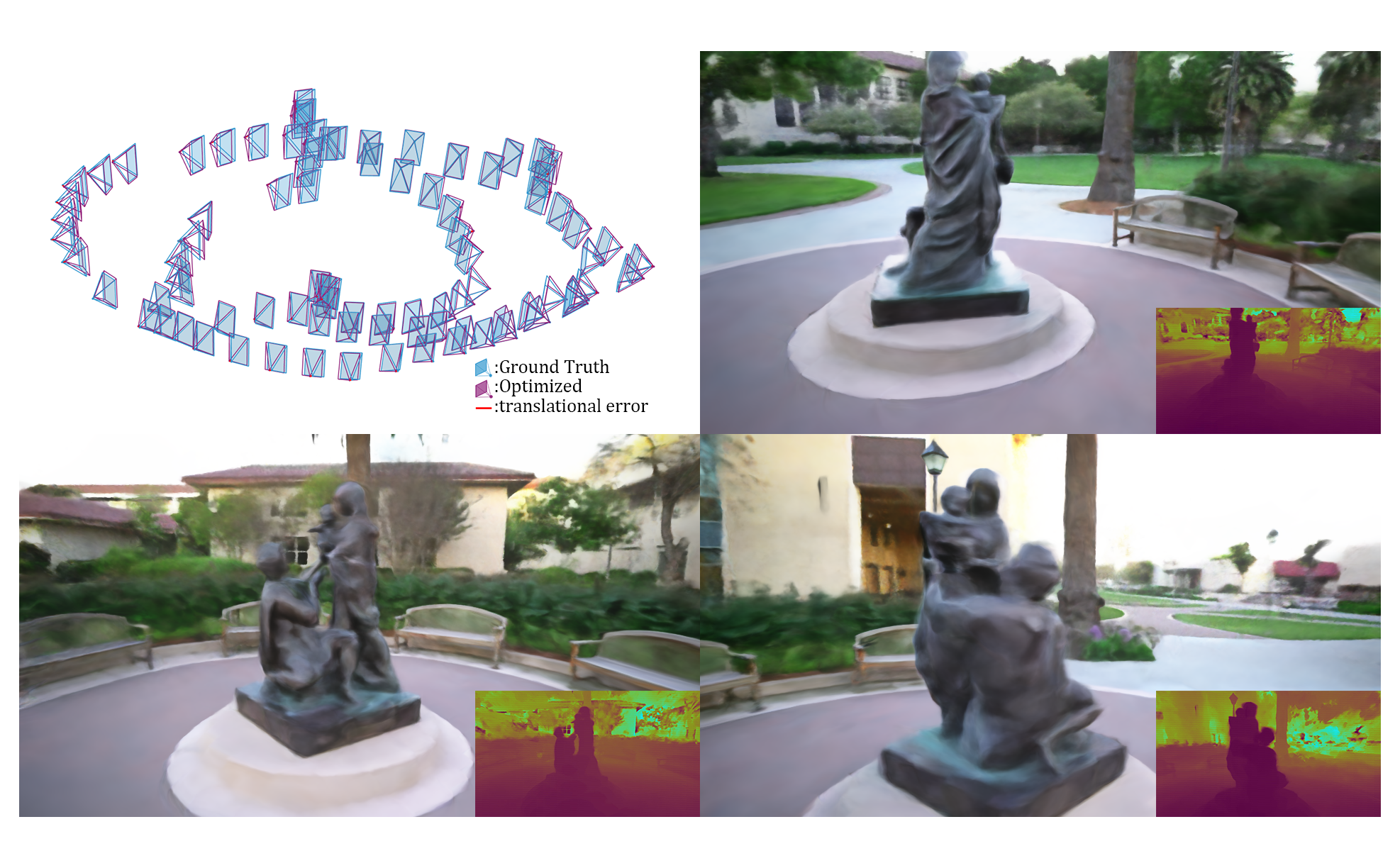}
    \caption{\textbf{Results of unordered sequence of images.} We extract these 100-frames from the scene of Family in Tanks and Temples dataset. The top left shows the pose distribution of the ground truth (COLMAP) and our optimized results. The translational errors are quite small for all trained frames. The rest of the images reveal the synthesis images and corresponding depth maps rendered from the trained NeRF model.}
    \label{fig:result-complex}
\end{figure}

In Tanks and Temples and CO3D V2 datasets, all frames are sampled from a continuous video, ensuring a strong spatial correlation between consecutive frames, and making it easier to determine their relative poses. Algorithms like NoPe-NeRF and CF-3DGS are designed based on this assumption, estimating relative poses under local adjacency or performing global incremental optimization. If the adjacency relationship in the input data is disrupted, these methods fail. Our method, however, is robust against such data, utilizing MST for association determination, a random matching mechanism in local optimization, and global optimization to achieve better results in these challenging scenarios. In the example shown in Fig.~\ref{fig:result-complex}, our method still demonstrates robust results.

\subsection{Ablation Study}

\textbf{Effectiveness of local/global joint optimization.}
To assess the effectiveness of our local joint phase, we remove it from the pipeline and begin global joint optimization directly from the training data preparation. Since we do not have a pre-trained rough NeRF model for evaluating the track points, we use projected 3D points from the mono-depth DPT as the initial track points. The results for novel view synthesis and pose accuracy are shown as \textit{w.o. local} in Tab.~\ref{table:ablation-local-global}. We also analyze the metrics of local-only results to highlight the need for global joint optimization, reporting these as \textit{w.o. global}. Our observations indicate that omitting either phase results in poorer metrics for view synthesis and pose estimation. Thanks to the estimated camera poses and the rough NeRF model from the local phase, along with pose refinement from the global phase, our joint optimization demonstrates outstanding performance in novel view synthesis and pose estimation.

\textbf{Comparison of NeRF with COLMAP poses.}
We further analyze the novel view synthesis quality of the NeRF model trained with our learned poses compared to COLMAP poses on the Tanks and Temples dataset. This evaluation allows us to indirectly compare the accuracy of the estimated poses based on their impact on novel view synthesis. Specifically, in Tab.~\ref{table:NeRF-COLMAP}, we achieve results that are comparable to COLMAP across all scenes, and in some cases, even outperform COLMAP. The main reason for this, as we analyzed, is that COLMAP uses traditional algorithms for pose estimation, which do not perform well in challenging scenarios with weak texture and large camera motions. In contrast, our proposed joint learning approach, through the design of various loss functions, not only ensures better rendering result quality but also enhances the accuracy of camera parameter estimation.

\textbf{Comparison to CF-3DGS with original 3DGS and our poses.}
To make the comparison of novel view synthesis with CF-3DGS reasonable, we also extract the estimated poses from the results of our final joint optimization and train an original 3DGS model with our poses. Tab.~\ref{table:ablation-3dgs-ours} shows the quantitative comparisons of 3DGS with our estimated camera poses and CF-3DGS on CO3D V2 dataset. Notice that based on the same 3DGS model, our results of novel view synthesis are much better than CF-3DGS, mainly benefiting from the more accurate pose estimation of our joint optimization, as exhibited in Tab.~\ref{table:pose&render-cf3dgs}.

\begin{table}[!t]
    \centering
    \resizebox{\linewidth}{!}{
    \begin{tabular}{c ccc c ccc c ccc}\hline
    \multirow{2}{*}{scenes}  & \multicolumn{3}{c}{w.o. local} & & \multicolumn{3}{c}{w.o. global} & & \multicolumn{3}{c}{Ours}\\ \cline{2-4}\cline{6-8}\cline{10-12}
                            & $RPE_t\downarrow$ & $RPE_r\downarrow$ & ATE$\downarrow$ & & $RPE_t$ & $RPE_r$ & ATE & & $RPE_t$ & $RPE_r$ & ATE  \\ \hline
    {110\_13051\_23361}     & 0.016 & 0.069 & \textbf{0.001}  & &  0.061& 0.253 & 0.005   & &  \textbf{0.015} &  \textbf{0.065} &   \textbf{0.001}  \\
    {415\_57112\_110099}    & 0.021 & 0.066 & \textbf{0.001}  & & 0.149 & 1.412 & 0.016   & &  \textbf{0.004} &  \textbf{0.020} &   \textbf{0.001}  \\
    {106\_12648\_23157}     & 0.034 & 0.325 & 0.003  & & 0.074 & 0.525   & 0.008 & &  \textbf{0.010} &  \textbf{0.116} &   \textbf{0.001}  \\
    {245\_26182\_52130}     & 0.012 & 0.066 & \textbf{0.001}  & & 0.056 & 0.324 & 0.006   & &  \textbf{0.011} &  \textbf{0.059} &   \textbf{0.001}  \\
    {34\_1403\_4393}        & 0.089 & 0.330 & 0.002  & & 0.108 & 0.390 & 0.010   & &  \textbf{0.030} & \textbf{0.110} & \textbf{0.001}  \\ \hline
    \end{tabular}
    }
    \caption{\textbf{Effectiveness of local and global joint optimization phase.}}
    \label{table:ablation-local-global}
\end{table}

\begin{table}[!t]
    \centering
    \resizebox{0.9\linewidth}{!}{\fontsize{10}{12}\selectfont
    \begin{tabular}{c ccc c ccc}\hline
    \multirow{2}{*}{scenes}  & \multicolumn{3}{c}{Ours} & & \multicolumn{3}{c}{NeRF+COLMAP} \\ \cline{2-4}\cline{6-8}
                            & PSNR$\uparrow$ & SSIM$\uparrow$ & LPIPS$\downarrow$ & & PSNR & SSIM & LPIPS\\ \hline
        Church    &  25.08 &  \textbf{0.76} &  \textbf{0.33}    &      & \textbf{25.72} &  0.75& 0.37\\
        Barn      &  26.69 &  \textbf{0.73} &  \textbf{0.39}    &      & \textbf{26.72} & 0.71 & 0.42 \\
        Museum    &  \textbf{27.55}&  \textbf{0.81}&  \textbf{0.29}&      & 27.21 & 0.78 & 0.34\\
        Family    &  \textbf{26.66}&  \textbf{0.78}&  \textbf{0.36}&      & 26.61 & 0.77 & 0.39 \\
        Horse     &  \textbf{27.08}&  \textbf{0.84}&  \textbf{0.25}&      & 27.02 & 0.82 & 0.29 \\
        Ballroom  &  \textbf{26.65}&  \textbf{0.81}&  \textbf{0.27}&      & 25.47 & 0.73 & 0.38 \\
        Francis   &  29.92&  \textbf{0.82}&  \textbf{0.36}&      & \textbf{30.05} & 0.81 & 0.38 \\
        Ignatius  &  \textbf{24.33} &  \textbf{0.63} &  \textbf{0.44}    &      & 24.08 & 0.61 & 0.47 \\ \hline
    \end{tabular}
    }
    \caption{\textbf{Comparison to NeRF trained with COLMAP poses.}}
    \label{table:NeRF-COLMAP}
\end{table}

\begin{table}[!t]
    \centering
    \resizebox{\linewidth}{!}{
    \begin{tabular}{c ccc c ccc}\hline
    \multirow{2}{*}{scenes}  & \multicolumn{3}{c}{3DGS+Ours} & & \multicolumn{3}{c}{CF-3DGS}\\ \cline{2-4}\cline{6-8}
                            & PSNR$\uparrow$ & SSIM$\uparrow$ & LPIPS$\downarrow$ & & PSNR & SSIM & LPIPS  \\ \hline
    {110\_13051\_23361}     & \textbf{31.60}& \textbf{0.93}&       \textbf{0.09}& & 29.69& 0.89& 0.29\\
    {415\_57112\_110099}    & \textbf{29.53}& \textbf{0.84}&       \textbf{0.13}& & 26.21& 0.73& 0.32\\
    {106\_12648\_23157}     & \textbf{26.13}& \textbf{0.83}&       \textbf{0.11}& & 22.14& 0.64& 0.34\\
    {245\_26182\_52130}     & \textbf{33.12}& \textbf{0.93}&       \textbf{0.14}& & 27.24& 0.85& 0.30\\
    {34\_1403\_4393}        & \textbf{31.33}& \textbf{0.94}&       \textbf{0.11}& & 27.75& 0.86& 0.20\\ \hline
    \end{tabular}
    }
    \caption{\textbf{Comparison to CF-3DGS with original 3DGS and our poses.}}
    \label{table:ablation-3dgs-ours}
\end{table}

\section{Conclusions}
\label{sec:conclusion}
In this work, we present NoPe-NeRF++, a novel local-to-global algorithm for joint camera pose estimation and NeRF training. Compared to previous methods, our approach can handle complex camera trajectories, dramatic camera viewpoint changes, and non-sequential image inputs. To tackle these challenges, our method begins by re-establishing the association relationships among input images through explicit feature matching for optimization. During the local joint optimization phase, constraints from local views, combined with a randomly selected non-adjacent reference image, enhance the robustness of camera poses and the NeRF model. Additionally, we introduce a global optimization phase that incorporates geometric consistency constraints via bundle adjustment, integrating feature trajectories to further improve pose estimation and the NeRF model. Extensive evaluations on benchmark datasets demonstrate our superior performance and robustness, even in challenging scenes. On the other hand, due to the limited expressive capability of NeRF itself, our method also has limitations in handling very large scenes, managing complex lighting conditions, and training/testing efficiency. In the future, we will consider more advanced representation methods, such as 3DGS~\cite{3DGS2023}, to address these issues. Because we only consider color and depth values of samples during the whole procedure, which are fundamental outputs of other representations. Therefore, it is intuitive to seamlessly integrate those models with our framework. Nevertheless, we believe our method represents an important step towards training NeRF models without pose priors for complex applications in the future.
\section{Acknowledgements}
\label{sec:acknowledgements}
We thank the anonymous reviewers for their constructive  comments that help improve this paper.

\printbibliography

\end{document}